\documentclass[10pt,twocolumn,letterpaper]{article}

\usepackage{cvpr}
\usepackage{times}
\usepackage{epsfig}
\usepackage{graphicx}
\usepackage{amsmath}
\usepackage{amssymb}
\usepackage{mathrsfs}
\usepackage{multirow}

\usepackage[pagebackref=true,breaklinks=true,letterpaper=true,colorlinks,bookmarks=false]{hyperref}

 \cvprfinalcopy % *** Uncomment this line for the final submission

\def\cvprPaperID{993} % *** Enter the CVPR Paper ID here

\ifcvprfinal\fi
\begin{document}

%%%%%%%%% TITLE
\title{Data Driven Robust Image Guided Depth Map Restoration}

\author{Wei Liu$^{1}$, Yun Gu$^{1}$, Chunhua Shen$^{2}$, Xiaogang Chen$^{3}$, Qiang Wu$^{4}$ and Jie Yang$^{1}$\\
$^1$Shanghai Jiao Tong University, Shanghai, China. {\tt\small \{liuwei.1989,geron762,jieyang\}@sjtu.edu.cn}\\
$^2$University of Adelaide, Adelaide, Australia. {\tt\small chunhua.shen@adelaide.edu.au}\\
$^3$University of Shanghai for Science and Technology, Shanghai, China. {\tt\small xg.chen@live.com}\\
$^4$University of Technology, Sydney, Australia. {\tt\small Qiang.Wu@uts.edu.au}
}

\maketitle
%\thispagestyle{empty}

%%%%%%%%% ABSTRACT
\begin{abstract}
Depth maps captured by modern depth cameras such as Kinect and Time-of-Flight (ToF) are usually contaminated by missing data, noises and suffer from being of low resolution. In this paper, we present a robust method for high-quality restoration of a degraded depth map with the guidance of the corresponding color image. We solve the problem in an energy optimization framework that consists of a novel robust data term and smoothness term. To accommodate not only the noise but also the inconsistency between depth discontinuities and the color edges, we model both the data term and smoothness term with a robust exponential error norm function. We propose to use Iteratively Re-weighted Least Squares (IRLS) methods for efficiently solving the resulting highly non-convex optimization problem. More importantly, we further develop a data-driven adaptive parameter selection scheme to properly determine the parameter in the model. We show that the proposed approach can preserve fine details and sharp depth discontinuities even for a large upsampling factor ($8\times$ for example). Experimental results on both simulated and real datasets demonstrate that the proposed method outperforms recent state-of-the-art methods in coping with the heavy noise, preserving sharp depth discontinuities and suppressing the texture copy artifacts.
\end{abstract}

%%%%%%%%% BODY TEXT
\section{Introduction}
\label{SecIntro}

%--------------------- Figure ----------------
\begin{figure}[t]
\centering
  \includegraphics[width=0.81\linewidth]{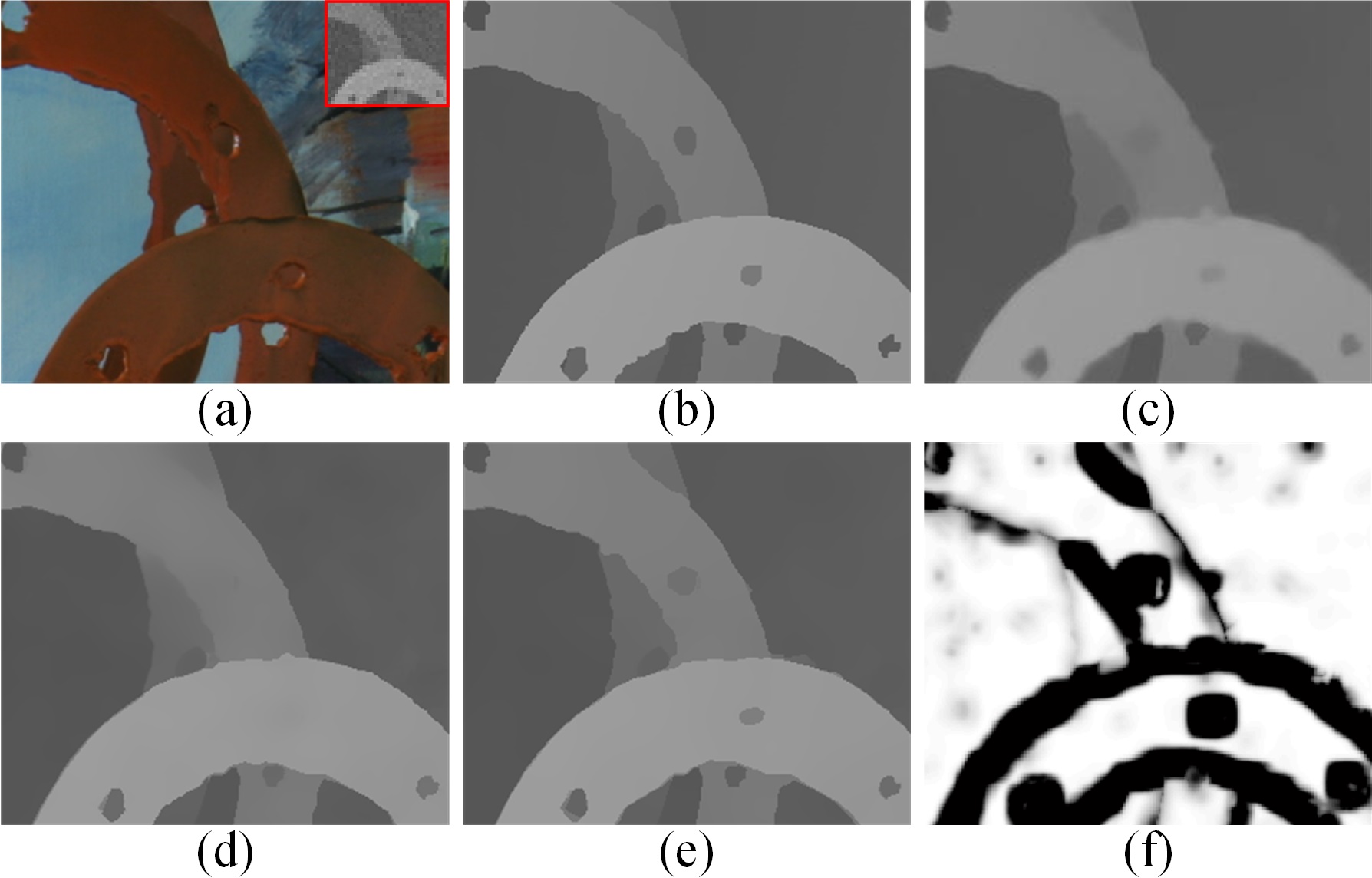}\\
  \caption{$8\times$ upsampling results. (a) The noisy low resolution depth map patch and the corresponding color image. (b) The ground truth. (c) The upsampling result of the state-of-the-art method of
  \cite{yang2014depth_recovery}. The upsampling result of our method (d) without adaptive bandwidth selection and (e) with adaptive bandwidth selection. (f) The corresponding bandwidth map of our adaptive bandwidth selection.}\label{FigCover}
\end{figure}
\vspace{-0.1cm}

%-----------------------------------------------

Acquisition of  depth information of 3D scenes is essential for many applications in computer vision and graphics. Applications range from 3D modeling to 3DTV and augmented reality.  A number of applications require accurate and high-resolution depth maps, for instance, object reconstruction, robot navigation and automotive driver assistance. Recently, modern depth cameras such as Kinect and Time-of-Flight (ToF) (e.g, {SwissRanger SR4000}) shown impressive results and become increasingly affordable. They can obtain dense depth measurements at a high frame rate. However, their depth maps usually suffer from missing values, noise and being of low resolution.

To facilitate the use of depth data, tremendous efforts have been spent on the restoration of  depth maps obtained by modern depth cameras. The depth map can be restored by different example-based methods such as
\cite{mac2012patch,li2014similarity,hornacek2013depth,ferstl2015variational}, which enhance the quality with a single depth map. This category of methods tend to fail to cope with large upsampling factors and most state-of-the-art methods mainly focus on up to $4\times$ upsampling. Another direction is to restore the depth map from multiple low-quality depth maps such as \cite{schuon2009lidarboost,hahne2011exposure,rajagopalan2008resolution,izadi2011kinectfusion}. This category of methods are more practical for static scenes than for dynamic environments. There are also strong research interests in developing image guided restoration schemes such as
\cite{diebel2005application,kopf2007joint,park2011high,yang2014depth_recovery,yang2007spatial,liu2013joint}, which restore the depth map with the guidance of the registered (aligned) color image. These methods are often based on the assumption that there exists a joint occurrence between depth discontinuities and color image edges. They can produce promising restoration quality with larger upsampling factors and are also not subject to static scenes when compared with the first two categories of methods. However, their depth maps often suffer from texture copy artifacts and blurring depth discontinuities when the depth discontinuities are inconsistent with the color edges.

In this paper, we propose a novel technique for robust image guide depth map restoration. The main contributions of our paper are as follows.
\begin{itemize}
  \itemsep -.4em
  \item[1.]
    To handle  heavy noises in  depth maps, we develop a robust data term that measures `pixel to patch' difference which is penalized by a robust error norm function within a Gaussian window. The proposed data term is inspired by the work in image denoising \cite{mrazek2006robust} and image editing \cite{an2008appprop}. We show that it is more robust in presence of heavy noises. To our knowledge, we are the first to introduce this robust data term in guided depth map restoration, and achieve impressive results.

  \item[2.] The proposed color image guided smoothness term is robust against the inconsistency between the depth discontinuities and the color edges. It has been proven to perform well in suppressing the texture copy artifacts and offer much better performance in preserving sharp depth discontinuities than the state-of-the-art methods such as the auto-regressive model \cite{yang2014depth_recovery}.

  \item[3.] While the proposed optimization framework is highly non-convex, we propose a numeric solution with an Iteratively Re-weighted Least Squares (IRLS) optimization framework, which can be efficiently solved and easily implemented by solving a linear system using off-the-shelf linear system solvers.

  \item[4.] More importantly, to eliminate heuristic parameter selection, we present a data-driven scheme to properly determine the parameter in our model such that fine details and sharp depth discontinuities are well preserved even for a large upsampling factor such as $8\times$, thus making the proposed method more practical.

 \end{itemize}

Experimental results on simulated and real data show that our method outperforms state-of-the-art methods in terms of both visual quality and accuracy.

%---------------------------------------
%\section
{\bf Related work and practical issues}
%\label{SecRelatedWork}
%
As our method belongs to the image guided restoration technique, we mainly review some recent guided restoration techniques and then the practical issues faced by this category of methods. Image guided depth restoration methods can be roughly classified as \emph{local methods} and \emph{global methods}. Local methods are based on a certain edge preserving filter where each pixel in the restored depth map is a weighted sum of its neighbors in the input depth map. The Joint Bilateral Upsampling (JBU) \cite{kopf2007joint} extended the bilateral filter \cite{tomasi1998bilateral} for depth upsampling where the bilateral weight is computed on the guidance color image. JBU was also used to fill the holes for Kinect depth map restoration in \cite{yang2012depth,yang2014depth_recovery}. To achieve sharper depth discontinuities, the work in \cite{liu2013joint} evaluated pixel dissimilarities based on the geodesic distance instead of the Euclidean distance in the JBU. Very recently, a new edge preserving filter named Guide Filter (GF) \cite{he2013guided} was introduced and used to perform joint upsampling.
Their work was further improved in \cite{lu2012cross,tan2014multipoint} to preserve sharp depth discontinuities and spatial variation. Global methods usually formulate the restoration as  optimization. The restored depth map is the minimal of the energy function. The authors of \cite{diebel2005application} performed restoration using an Markov Random Field (MRF) formulation with a pairwise appearance consistency data term and an image guided smoothness term. Their work was further extended in \cite{park2011high} by incorporating different weighting schemes to combine different cues including segmentation, image gradients, edge saliency and a non-local means. The work in \cite{yang2012depth,yang2014depth_recovery} performed image guided depth map restoration using an color guided auto-regressive model. Sparse representation model \cite{kiechle2013joint,kwon2015data} learns statistical dependencies between intensity and depth in a scene.

The fundamental assumption of image guided depth map restoration is that there exists a joint occurrence between depth discontinuities and color image edges. However, when the depth discontinuities are inconsistent with the color edges, these methods may face the following two problems: 1) texture copy artifacts on the smooth depth region when the corresponding color image is highly textured; 2) blurring depth discontinuity when the corresponding color image is more homogeneous.  To handle  these two issues, most recent work proposed to take the bicubic interpolation of the noisy low resolution depth map into account such as the definition of the auto-regressive coefficient in \cite{yang2014depth_recovery} and the RGB-D structure similarity in \cite{kwon2015data}. However, the bicubic interpolation of the input depth map becomes unreliable especially when the upsampling factor is large (e.g., $8\times$) and the input depth map contains heavy noises. This situation is often the case as the resolution of modern depth cameras (e.g., {SwissRanger SR4000}) is typically low and also contains heavy noises. We attempt to  alleviate these two issues in this paper. The numeric solution of the proposed model shows that our model can iteratively exploit the newly updated depth map at each iteration. The quality of the newly updated depth map is considerably better than the bicubic interpolation of the input depth map. We show that this helps to preserve much sharper depth discontinuities than the previous work such as the auto-regressive model \cite{yang2014depth_recovery}. Also, how to well handle the heavy noises in the depth map is another important issue. Nevertheless, little work has been done on this issue. Here we demonstrate the robustness of our energy minimization model, due to the newly proposed robust data term.

%---------------------------------------
\section{The proposed method}
\label{SecTheMethod}

%----------------------------
{\bf The model}
Our upsampling model consists of two terms: the data term and the smoothness term. Given a noisy low resolution depth map $D_L$, it is first interpolated to $D^0$ by bicubic interpolation. $D^0$ has the same resolution as the guidance image. Then our upsampling model is formulated as:
\begin{equation}\label{EqMyModel}
\small
    D_H=\underset{D}{\mathop{\arg \min}}\,\left\{( 1-\alpha){{E}_{D}}(D,{{D}^{0}})+\alpha {{E}_{S}}(D)\right\}
\end{equation}
where ${{E}_{D}}\left( D,{{D}^{0}} \right)$ is the data term that makes the result to be consistent with the input. ${{E}_{S}}\left( D \right)$ is the smoothness term that reflects prior knowledge of the smoothness of our solution. The relative importance of these two terms is balanced with the parameter $\alpha $.\\
\textbf{The data term ${{E}_{D}}\left( D,{{D}^{0}} \right)$}: the data term is defined as:
\begin{equation}\label{EqDataTerm}
\small
    {{E}_{D}}\left( D,{{D}^{0}} \right)=\sum\limits_{i\in \Omega }{\sum\limits_{j\in N\left( i \right)}{{{\omega }_{i,j}}{{\varphi }_{D}}(|{{D}_{i}}-D_{j}^{0}|^2)}}
\end{equation}
where $\Omega$ represents the set of the high resolution coordinates. $N(i)$ is the neighborhood of $i$, namely, the square patch of radius $r$ centered at $i$. The Gaussian window $\omega_{i,j}$ decreases the weights when $j$ is far from $i$:
\begin{equation}\label{EqSpatialWeight}
\small
    {{\omega }_{i,j}}=\exp \left( -\frac{|i-j{{|}^{2}}}{2\sigma _{s}^{2}} \right)
\end{equation}
where ${{\sigma }_{s}}$ is a constant that is defined by the user. ${\varphi}_{D}(\cdot)$ is the robust error norm function that we denote as the exponential error norm:
\begin{equation}\label{EqErroNormFunction}
\small
    {{\varphi}_{D}}(x^2)=2{{\lambda }^{2}}\left( 1-\exp \left( -\frac{{{x}^{2}}}{2{{\lambda }^{2}}} \right) \right)
\end{equation}
where $\lambda$ is a user defined constant.

The proposed data term is inspired by the work in image denoising \cite{mrazek2006robust} and image editing \cite{an2008appprop}. Their data term replaces the 'pixel to pixel' difference with 'pixel to patch' difference and has been shown robust against noisy input images (for image denoising \cite{mrazek2006robust}) and inaccurate input strokes (for image editing \cite{an2008appprop}). As depth maps obtained by a modern depth camera such as {SwissRanger SR4000} are usually of low resolution and contaminated by heavy noises, we thus also employ the similar 'pixel to patch' difference measurement in the data term. The 'pixel to patch' difference may blur the depth discontinuities for the pixels around the depth discontinuities. To better preserve the depth discontinuities, we adopt the robust error norm function defined in Eq.~(\ref{EqErroNormFunction}) to model the data term instead of the  $L_2$ error norm. This is because the error norm function used here is known to be robust against the outliers and thus could better preserve the depth discontinuities.
The Gaussian window is introduced to further reduce the influence of the pixels far from the central pixel.

Our data term is different from that of previous methods \cite{diebel2005application,park2011high,yang2014depth_recovery}
that only use 'pixel to pixel' difference modeled by the $L_2$ error norm. Their data term works well for input depth maps with high accuracy, but fails when heavy noises are presented in the data. In fact, our data term can be viewed as a generalized form of their data term. If we set the radius of the neighborhood $N(i)$ of pixel $i$ as $r=0$, then our data term  becomes to the conventional non-robust form.\\
\textbf{The smoothness term ${{E}_{S}}(D)$}: Our smoothness is guided by the aligned color image. It is defined as:
\begin{equation}\label{EqSmoothnessTerm}
\small
E_S(D) =\sum\limits_{i\in \Omega}\sum\limits_{j\in N(i)}\omega_{i,j}^c\varphi_S\left(|D_i-D_j|^2\right).
\end{equation}

The guide weight $\omega_{i,j}^c$ is defined as:
\begin{equation}\label{EqColorSpatialWeight}
\small
    \omega_{i, j}^c=\omega_{i,j}\cdot \exp\left(-\frac{\sum\limits_{k\in C}{|I_{i}^{k}-I_{j}^{k}{{|}^{2}}}}{3\times2\sigma _{c}^{2}}\right)
\end{equation}
where $\omega_{i,j}$ is the same as Eq.~(\ref{EqSpatialWeight}). $C=\left\{ R,G,B \right\}$ represents the different channels of the color image. ${{\sigma }_{c}}$ is a constant defined by the user. In fact, there are also other choice of the guide weight such as the shape-based structure aware weight in \cite{yang2014depth_recovery} or the more complex guide weight in \cite{park2011high}. However, we find that the adopted guide weight has already been enough to yield satisfying results. To adopt the weight in \cite{yang2014depth_recovery} and \cite{park2011high} shows little performance improvement while their computational cost is much higher than Eq.~(\ref{EqColorSpatialWeight}).

We also employ the function in Eq.~(\ref{EqErroNormFunction}) to model the smoothness term, i.e.:
\begin{equation}\label{EqSoomthnessNormFun}
  {\varphi }_S(\cdot)={\varphi}_{D}(\cdot).
\end{equation}

As we will show in the sequel, the numeric solution to our model shows that the employed $\varphi_S(\cdot)$ makes the model robust against the inconsistency between the depth discontinuities and the color edges. This is significant in suppressing the texture copy artifacts and preserve sharp depth discontinuities.

%------------------------------------
\subsection{The numeric solution and analysis}
\label{SecAppSolutionFurAnalysis}

In this section, we first present the numeric solution to our formulated problem. Then we show further analysis why our model can handle the three practical issues discussed above, namely, the heavy noise in the input depth map, the texture copy artifacts and the blurring depth discontinuities caused by the inconsistency between the depth discontinuities and the color edges.

Due to the highly non-convex property of the proposed model, directly solving it is challenging. Previous work such as the one in \cite{lu2011revisit} used the Loop Belief Propagation (LBP) \cite{yedidia2000generalized} to solve their
energy minimization function. However, classical energy minimization solvers such as the LBP \cite{yedidia2000generalized} and graph cuts \cite{boykov2001fast,kolmogorov2004energy} work for discrete energy minimization. In our problem, the variables are naturally real-valued and one has to discretize them in order to apply LBP or graph cuts. The computational cost can be extremely expensive when the continuous problem is discretized into $8192$ levels ($13$ bits). This is the practical issue as modern depth cameras usually save the depth map in a $16$ bits image (with $13$ bits for depth values). The quantization error can be large if only a few quantization levels are used. In contrast, we present a numeric solution to our model that works for continuous system and can be efficiently solved. First, we present the normal equation of our model:
\begin{equation}\label{EqNormalEquation}
\footnotesize
\begin{split}
    &\frac{\partial{E}}{\partial{D_i}}=(1-\alpha)\sum\limits_{j\in N(i)}\omega_{i,j}d_{i,j}\left(D_i-D_j^0\right)\\
    &\ \ \ \ \ \ \ \ \ \ \ \ \ +2\alpha\sum\limits_{j\in N(i)}\omega_{i,j}^cs_{i,j}\left(D_i-D_j\right) = 0, \ i \in \Omega\\
\end{split}
\end{equation}
where we define
\begin{equation}\label{EqErrorNormFunctionDerivative}
\small
\begin{split}
    &d_{i,j}={\varphi}'_{D}(|D_i - D^0_j|^2), \ \ \ s_{i,j}={\varphi}'_{S}(|D_i - D_j|^2),\\
    &\ \ \ \ \ \ \ \ \ \ \ \ \ {\varphi}'_{D}(x^2)={\varphi}'_{S}(x^2)=\exp\left(-\frac{x^2}{2\lambda^2}\right)
\end{split}
\end{equation}
${{{\varphi }'}_{D}}(x^2)={{{\varphi }'}_{S}}\left(x^2 \right)$ is the derivative of ${{\varphi }_{D}}\left( x^2 \right)={{\varphi }_{S}}\left( x^2 \right)$ defined in Eq.~(\ref{EqErroNormFunction}).
%

%----------------------------- Figure -------------------------------
\begin{figure}
\centering
  \includegraphics[width=1\linewidth]{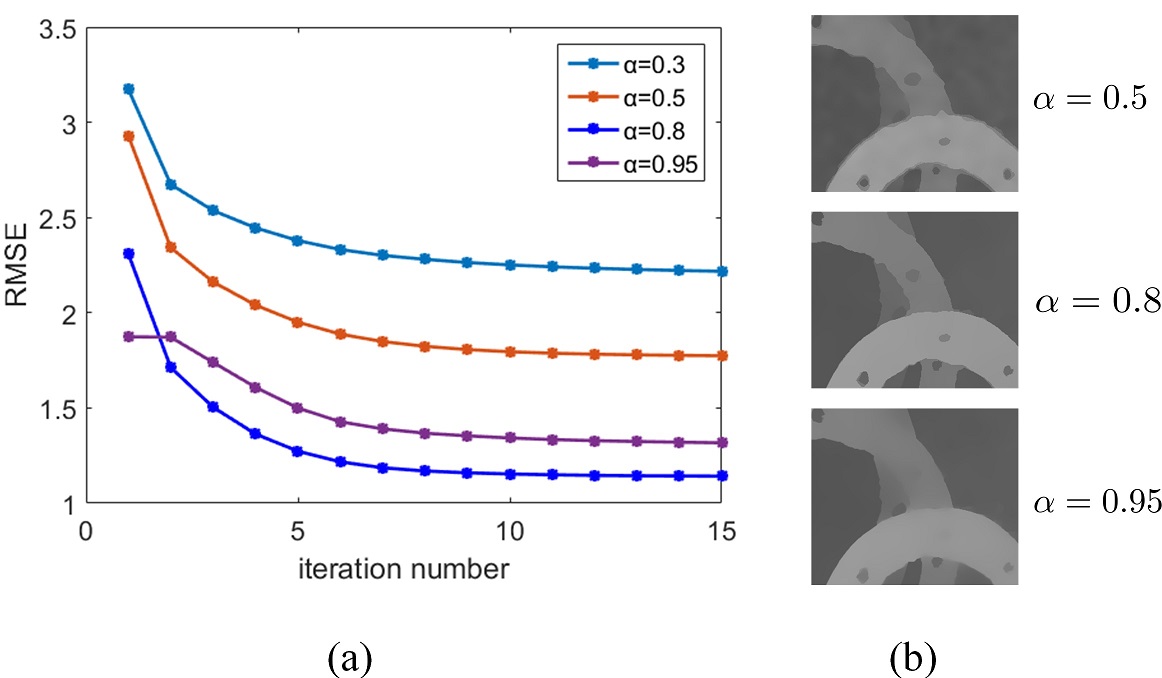}\\
  \caption{(a) Convergency analysis of the proposed model on $8\times$ upsampling for different $\alpha$ values in terms of RMSE between the newly updated depth map and the groundtruth in each iteration. (b) Example results of different $\alpha$ values.}\label{FigCovergeAnalysis}
\end{figure}
%------------------------------------------------

A closed-form solution to Eq.~(\ref{EqNormalEquation}) is not available, we can only solve it iteratively. If we keep $s_{i,j}, d_{i,j}$ as constant in each iteration where $d_{i,j}^n={\varphi}'_{D}(|D_i^n - D^0_j|^2), s_{i,j}^n={\varphi}'_{D}(|D_i^n - D^n_j|^2)$ for iteration $n+1$, then Eq.~(\ref{EqNormalEquation}) becomes the standard form of the  following re-weighted least squares optimization framework as:
\begin{equation}\label{EqIterReWeightLeastSquare}
\small
\begin{split}
    &D^{n+1}=\underset{D}{\arg \min}\{(1-\alpha)\sum\limits_{i\in\Omega}\sum\limits_{j\in N(i)}\omega_{i,j}d_{i,j}^n(D_i-D_j^0)^2 \\
    &\ \ \ \ \ \ \ \ \ \ \ \ \ \ \ \ \ \ \ \ \ \ \ \ \ \ \ \ \ \ \ \ \ \ \ \ + \alpha\sum\limits_{i\in\Omega}\sum\limits_{j\in N(i)}\omega_{i,j}^{c}s_{i,j}^n(D_i-D_j)^2\}
\end{split}
\end{equation}

Then we can iteratively solve Eq.~(\ref{EqIterReWeightLeastSquare}) until the final output meets the convergence condition. This approximation is  similar to the well-known Iteratively Re-weighted Least Squares (IRLS) \cite{chartrand2008iteratively} in the literature. However, their IRLS is only suitable for $L_p (0<p<2)$ norm optimization framework. In this paper, we also denote Eq.~(\ref{EqIterReWeightLeastSquare}) as IRLS. As Eq.~(\ref{EqIterReWeightLeastSquare}) is quadratic in each iteration, thus it can be minimized by solving the set of linear equations:
\begin{equation}\label{EqLinearSystem}
\footnotesize
\begin{split}
   &\left[(1-\alpha)\sum\limits_{j\in N(i)}\omega_{i,j}d_{i,j}^n+2\alpha\sum\limits_{j\in N(i)}\omega_{i,j}^{c}s_{i,j}^n\right]D_i\\
   &-2\alpha\sum\limits_{j\in N(i)}\omega_{i,j}^{c}s_{i,j}^nD_j=(1-\alpha)\sum\limits_{j\in N(i)}\omega_{i,j}d_{i,j}^nD_j^0
\end{split}
\end{equation}
we rewrite Eq.~(\ref{EqLinearSystem}) in matrix notation as:
\begin{equation}\label{EqLinearSystemMatrix}
\small
\begin{split}
    &\ \ \ \ \ \ \ \ \ \left[(1-\alpha)W^n-2\alpha S^n\right]D = (1-\alpha)Z^nD^0\\
    &\Longrightarrow D^{n + 1}=(1-\alpha)\left[(1-\alpha)W^n-2\alpha S^n\right]^{-1}Z^nD^0
\end{split}
\end{equation}
where $D$ and $D^0$ are the vectors of the updated depth map and the initial depth map respectively. We use bicubic interpolation of the input depth map as initialization. $W^n$ is a diagonal matrix with $W^n_{i,i}=\sum\limits_{j\in N(i)}\omega_{i,j}d^n_{i,j}+\frac{2\alpha}{1-\alpha}\omega^c_{i,j}s^n_{i,j}$. $S^n$ is the affinity matrix whose elements are $S^n_{i,j}=\omega^c_{i,j}s^n_{i,j}$. $Z^n$ is another affinity matrix with $Z^n_{i,j} = \omega_{i,j}d^n_{i,j}$.

%----------------------------- Figure -------------------------------
\begin{figure}
\centering
  \includegraphics[width=1\linewidth]{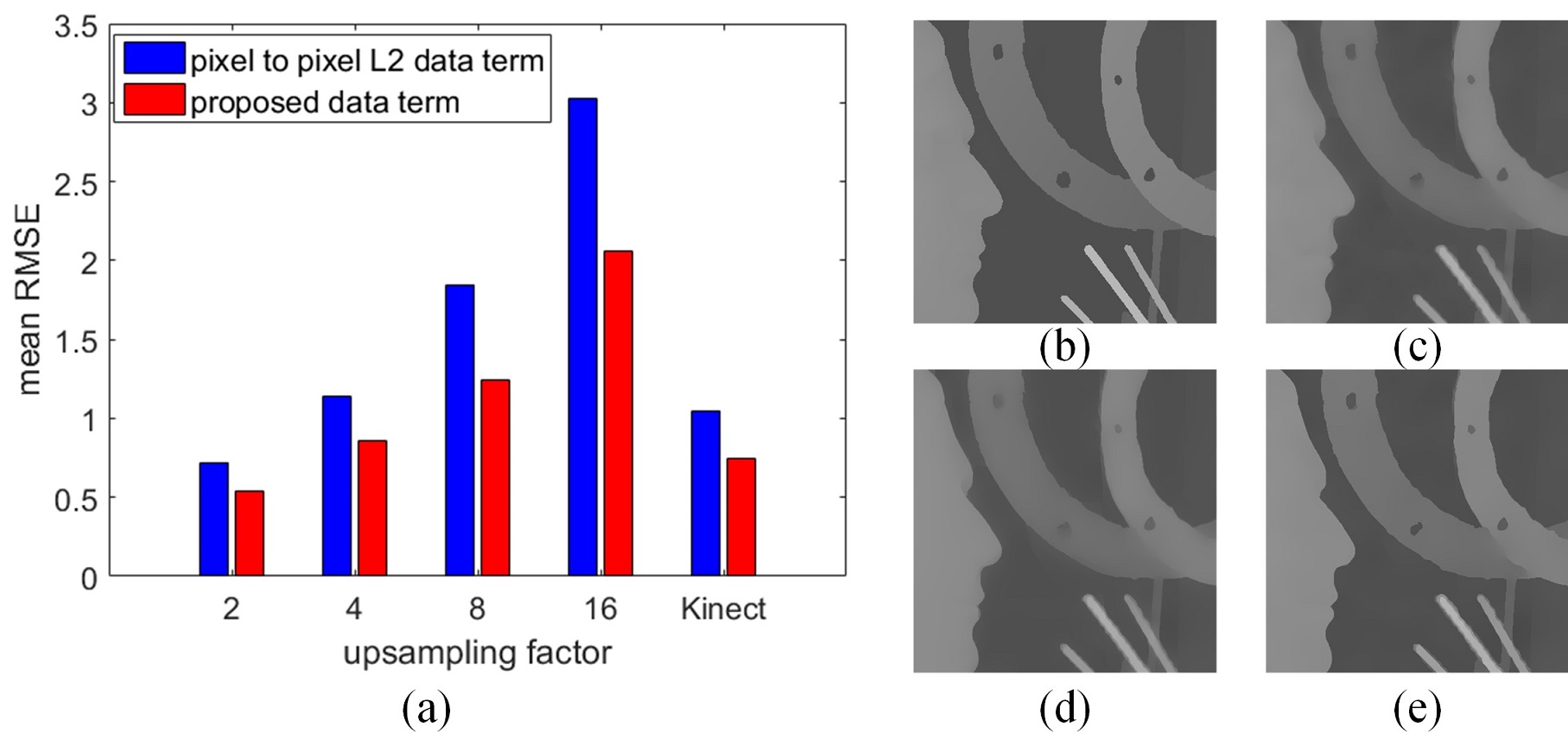}\\
  \caption{(a) Quantitative comparison of the proposed data term vs.\
  the `pixel to pixel' difference $L_2$ error norm data term in terms of RMSE on different upsampling factors of simulated ToF data \cite{yang2014depth_recovery} upsampling and simulated Kinect data \cite{lu2014depth} restoration. Examples of (b) the groundtruth, (c) results of pixel to pixel difference $L_2$ error norm data term, (d) first denoise the depth map with BM3D \cite{dabov2007image} and then upsample the depth map with the pixel to pixel difference $L_2$ error norm data term, (e) results of the proposed data term. }\label{FigDifferentDataTermComp}
\end{figure}
%------------------------------------------------
%------------------------------------- Figure -----------------------------
\begin{figure*}
\centering
  \includegraphics[width=0.75\linewidth]{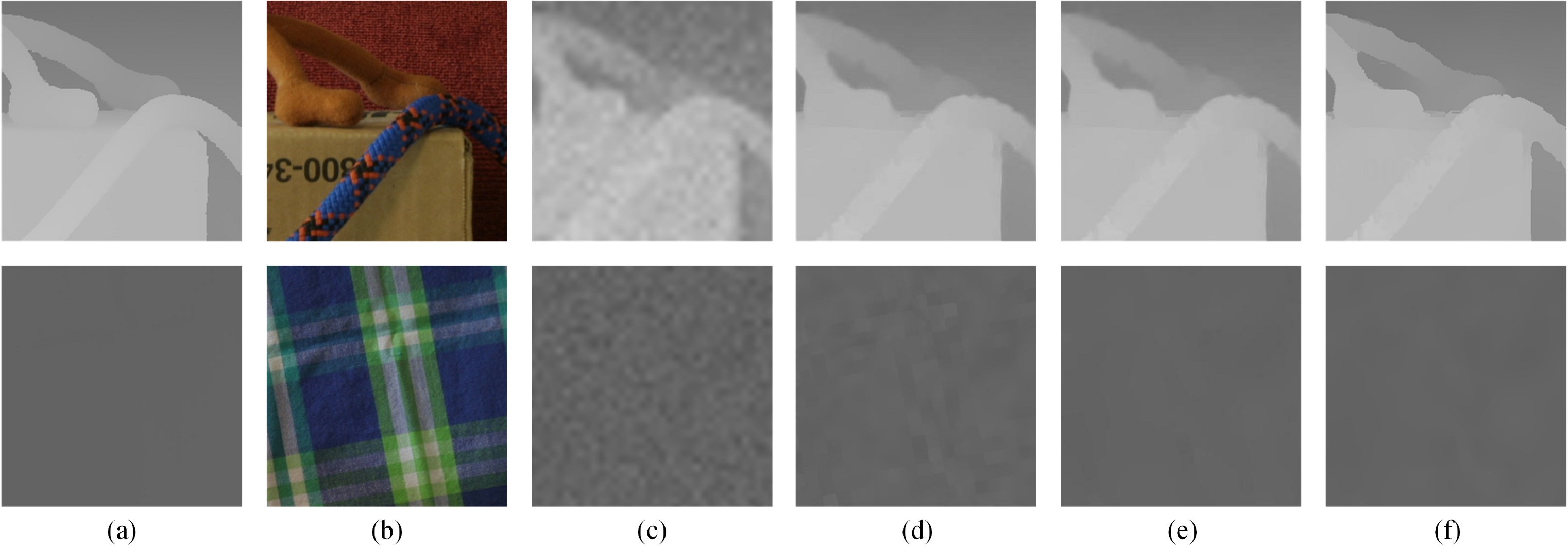}\\
  \caption{Comparison of depth discontinuities preserving and texture copy artifacts suppression on $8\times$ upsampling. (a) The groundtruth. (b) The guidance color images. (c) The bicubic interpolation of the input depth map. Results obtained by the MRF \cite{diebel2005application}, (e) the auto-regressive model \cite{yang2014depth_recovery} and (f) our method. The first row shows depth discontinuities preserving comparison. The second row shows texture copy artifacts suppression comparison.}\label{FigMotivation}
\end{figure*}
%------------------------------------------------------------------------

Solving the linear equation in Eq.~(\ref{EqLinearSystemMatrix}) has long been studied in the literature and there are many efficient modern solvers. In our experiment, we use the Preconditioned Conjugate Gradient (PCG) in \cite{krishnan2013efficient} to solve Eq.~(\ref{EqLinearSystemMatrix}) which shows to be very efficient and can produce good results. Besides the efficiency, our IRLS also has good convergence property. Fig.~\ref{FigCovergeAnalysis}(a) shows the convergence analysis for different $\alpha$ values. It is clear that our IRLS shows good convergence property for a large range of $\alpha$ values. Experiments on other upsampling factors and the Kinect dataset show  similar performance. Note that the $\alpha$ value used in the our experiments is based on the  consideration of noise smoothing and fine details preserving. Small $\alpha$ cannot well smooth the noise while large $\alpha$ can over smooth fine details. Fig.~\ref{FigCovergeAnalysis}(b) illustrates three example results of different $\alpha$ values.

Now we show how the proposed method may alleviate the three practical issues mentioned at the beginning of this section. First, note that on the right of Eq.~(\ref{EqLinearSystem}) is $\sum\limits_{j\in N(i)}\omega_{i,j}d_{i,j}^nD_j^0$ for each pixel $i\in \Omega$. This is in fact the filtered output of the initial depth map $D^0$ which results from our novel data term. If we use the data term that measures the $L_2$ norm of the pixel to pixel difference, then the right side of Eq.~(\ref{EqLinearSystem}) will be only $D^0_i$ for each pixel $i\in \Omega$. This will be quite noisy for noisy input and large upsampling factor. On the contrary, the filtering on the right side of Eq.~(\ref{EqLinearSystem}) can well smooth the noise in $D^0$. This makes our model more robust against the noise. To further validate our analysis, we replace the proposed data term in our model with the pixel to pixel difference measurement $L_2$ error norm data term and perform experiments on both the simulated ToF dataset \cite{yang2014depth_recovery} and Kinect dataset \cite{lu2014depth}. Then we compare the mean RMSE of the results. Fig.~\ref{FigDifferentDataTermComp} shows the comparison results. As shown in the figure, the proposed data term can clearly improve the performance especially for large upsampling factors. Fig.~\ref{FigDifferentDataTermComp}(c) and (e) also illustrate examples for visual comparison where the proposed data term also shows better visual quality.

In fact, the proposed data term is equivalent to the $L_2$ norm `pixel to pixel' data term of which the input depth map is a filtered depth map as the right side of Eq.~(\ref{EqLinearSystem}). However, this is different from the way that one firstly denoises the input depth map with a  denoising method such as \cite{dabov2007image} and then performs upsampling with the $L_2$ norm `pixel to pixel' data term. This is because our filtering weights $d_{i,j}^n$ is based on the newly updated depth map. Moreover, as pointed out in \cite{mac2012patch}, first denoising the depth map followed by upsampling can destroy small structures in the depth map, especially for large upsampling factors. This is also validated in our experiments as illustrated in Fig.~\ref{FigDifferentDataTermComp}(d).

To handle the inconsistency between the depth discontinuities and the color edges, most recent work proposed to take the bicubic interpolation of the input depth map into account \cite{yang2014depth_recovery,kwon2015data}. This can efficiently suppress the texture copy artifacts but fail to properly preserve sharp depth discontinuities when the upsampling factor is large and the input depth map contains heavy noise. This is because the depth discontinuities have already been  blurred for the bicubic interpolation of large upsampling factor as can be seen in Fig.~\ref{FigMotivation}(c). Note the guide weight of the smoothness term of our IRLS in Eq.~(\ref{EqIterReWeightLeastSquare}) $s_{i,j}^n$ is based on the newly updated depth map of which the quality is much better than the bicubic interpolation. Fig.~\ref{FigMotivation} shows the comparison of three different methods: 1) the Markov Random Field (MRF) in \cite{diebel2005application} which only use the color information for smoothness term weights, 2) the auto-regressive model in \cite{yang2014depth_recovery} which use both the color information and the bicubic interpolation of the input depth map, 3) our method which use both the color information and the newly updated depth map. It is clear that our method can not only suppress the texture copy artifacts but also preserve sharper depth discontinuities than the auto-regressive model in \cite{yang2014depth_recovery}.

%-------------------------------------
\subsection{Data driven parameter selection}
\label{SecBandwidthSelection}

The parameter $\lambda $ in Eq.~(\ref{EqErroNormFunction}) is an important parameter in our model. A large $\lambda $ will have better noise smoothing but may over smooth the depth discontinuities. Thus, pixel in the homogeneous regions on the depth map should be assigned with large $\lambda$. A small $\lambda $ could better preserve the depth discontinuities but performs poorly in noise smoothing. Thus, pixels along the depth discontinuities should be assigned with small $\lambda$. We denote $\lambda$ as the bandwidth of the exponential error norm function. To eliminate heuristic parameter selection, we describe another optimization framework that adapts $\lambda $ to each pixel in the depth map resulting in a data driven adaptive parameter selection model. Because the depth map is quite piece wise smooth, we assume that the bandwidth is also regular and smooth. Therefore, we add another term that consists of the $L_2$ norm of the gradient of ${\lambda}_{i}(i\in\Omega)$  to the objective function in Eq.~(\ref{EqMyModel}), resulting in the following objective function:
\begin{equation}\label{EqBandwidthModel}
\small
    E=\left( 1-\alpha  \right){{E}_{D}}\left( D,{{D}^{0}} \right)+\alpha {{E}_{S}}\left( D \right)+\beta \sum\limits_{i\in \Omega }{|\nabla {{\lambda }_{i}}{{|}^{2}}},\ i\in \Omega
\end{equation}
By minimizing Eq.~(\ref{EqBandwidthModel}) with respect to ${{\lambda }_{i}}$ through the steepest gradient descent according to:
\begin{equation}\label{EqBandwidthUpdate}
\small
    \lambda _{i}^{n+1}=\lambda _{i}^{n}-\tau \frac{\partial E}{\partial \lambda _{i}^{n}},\ i\in \Omega
\end{equation}
where $\tau $ is the given updating rate and the derivative of the objective function is given by
\begin{equation}\label{EqBandwidthDerivative}
\footnotesize
\begin{split}
    &\frac{\partial E}{\partial \lambda^n_i} = \\
    &( 1-\alpha)\sum\limits_{j\in N(i)}{{{\omega }_{i,j}}\left[ 4{{\lambda }_{i}^n}\left( 1-{{d}_{i,j}^n} \right)-\frac{2{{\left( {{D}^n_{i}}-D_{j}^{0} \right)}^{2}}}{{{\lambda}_{i}^n}}{{d}^n_{i,j}} \right]}+\\
    &\alpha \sum\limits_{j\in N\left( i \right)}{{{{\tilde{\omega }}}_{i,j}}\left[ 4{{\lambda}_{i}^n}\left( 1-{{s}^n_{i,j}} \right)-\frac{2{{\left( {{D}^n_{i}}-{{D}^n_{j}} \right)}^{2}}}{{{\lambda }_{i}^n}}{{s}^n_{i,j}} \right]}+2\beta \sum\limits_{i\in \Omega }{\Delta {{\lambda }_{i}^n}}
\end{split}
\end{equation}
where ${{d}_{i,j}^n}$ and ${{s}_{i,j}^n}$ are the same as Eq.~(\ref{EqErrorNormFunctionDerivative}).

%----------------------------- Figure -------------------------------
\begin{figure}
\centering
  \includegraphics[width=1\linewidth]{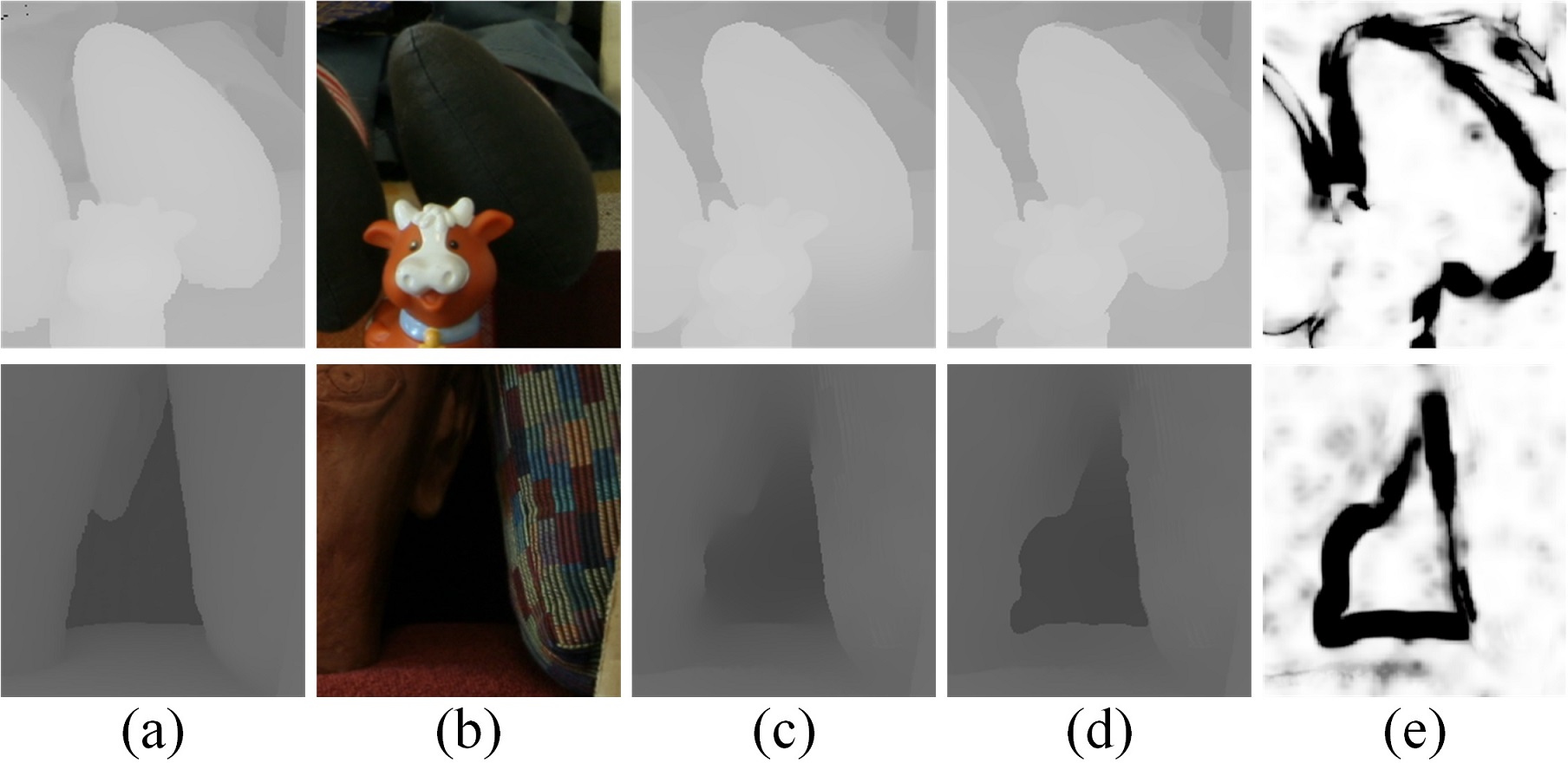}\\
  \caption{Visual comparison of our method for $8\times$ upsampling with and without bandwidth selection. (a) The groundtruth. (b) The corresponding color image. (c) Results obtained without bandwidth selection. (d) Results obtained with bandwidth selection and (e) the corresponding bandwidth maps.}\label{FigParaSelectAdvantage}
\end{figure}
%------------------------------------------------

In our experiments, depth map updating and the parameter selection are addressed in an iterative fashion through alternating the parameter update in Eq.~(\ref{EqBandwidthUpdate}) and the depth map update in Eq.~(\ref{EqLinearSystemMatrix}). Fig.~\ref{FigCover}(f) illustrates a bandwidth map obtained by our method. It is clear that this bandwidth map well corresponds to the character of the depth map shown in Fig.~\ref{FigCover}(b). The adaptive selection around depth discontinuities corresponds to lower values than smooth homogeneous regions. Visual comparison illustrated in Fig.~\ref{FigCover}(d) and (e) shows that the proposed bandwidth selection helps to preserve fine details even for the $8\times$ upsampling. Fig.~\ref{FigParaSelectAdvantage} further shows that the proposed bandwidth selection can also help to preserve sharp depth discontinuities even for $8\times$ upsampling.

%------------------------------
\section{Experiments}
\label{SecExperiment}

In this section, we compare our method with state-of-the-art methods for depth map restoration using the simulated ToF dataset from \cite{yang2014depth_recovery}, simulated Kinect dataset from \cite{lu2014depth}, the real ToF dataset from \cite{ferstl2013image} and the NYU Kinect dataset \cite{silberman2012indoor}. All the experimental results in this paper and additional experimental results, including qualitative comparisons using other Kinect scene datasets, are provided in the supplemental material.

%--------------------------------------- Table ---------------------------------
\begin{table*}
\centering
\caption{Quantitative comparison on the simulated ToF data using the noisy Middlebury dataset from \cite{yang2014depth_recovery}. Results are evaluated in RMSE and the best results are in bold.}\label{TabSimulatedRMSE}

\resizebox{1\textwidth}{!}
{
\begin{tabular}{|c|cccc|cccc|cccc|cccc|cccc|cccc|}

\hline
  \multicolumn{1}{|c}{\multirow{2}{*}{}} & \multicolumn{4}{|c|}{\emph{Art}} & \multicolumn{4}{c|}{\emph{Book}} & \multicolumn{4}{c|}{\emph{Dolls}} & \multicolumn{4}{c|}{\emph{Laundry} } & \multicolumn{4}{c|}{\emph{Moebius}} & \multicolumn{4}{c|}{\emph{Reindeer}}\\

  \cline{2-25} %\cline{6-9} \cline{10-13} \cline{14-17} \cline{18-21} \cline{22-25}
  & $2\times$ & $4\times$ & $8\times$ & $16\times$ & $2\times$ & $4\times$ & $8\times$ & $16\times$ & $2\times$ & $4\times$ & $8\times$ & $16\times$ & $2\times$ & $4\times$ & $8\times$ & $16\times$ & $2\times$ & $4\times$ & $8\times$ & $16\times$ & $2\times$ & $4\times$ & $8\times$ & $16\times$ \\
  \hline

  JGF \cite{liu2013joint} & 2.36 & 2.74 &	3.64 & 5.46 & 2.12 & 2.25 &	2.49 & 3.25 & 2.09 & 2.24 &	2.56 & 3.28 & 2.18 & 2.4 & 2.89 & 3.94 & 2.16 & 2.37 & 2.85 &	3.9 & 2.09 & 2.22 &	2.49 & 3.25 \\

  NLM-MRF \cite{park2011high} & 1.66 & 2.47 & 3.44 & 5.55 & 1.19 & 1.47 & 2.06 & 3.1 & 1.19 & 1.56 & 2.15 & 3.04 & 1.34 & 1.73 & 2.41 & 3.85 & 1.2 & 1.5 & 2.13 & 2.95 & 1.26 & 1.65 & 2.46 & 3.66 \\

  JID \cite{kiechle2013joint} & 1.69 & 2.98 & 3.68 & 5.99 & 1.53 & 2.71 & 3.04 & 4.39 & 1.54 & 2.71 & 2.94 & 3.90 & 1.45 & 2.72 & 3.16 & 4.63 & 1.55 & 2.72 & 2.94 & 4.34 & 1.65 & 2.80 & 3.13 & 4.63 \\

  TGV \cite{ferstl2013image} & 0.8 & 1.21 & 2.01 & 4.59 & 0.61 & 0.88 & 1.21 & 2.19 & 0.66 & 0.96 & 1.38 & 2.88 & 0.61 & 0.87 & 1.36 & 3.06 & 0.57 & 0.77 & 1.23 & 2.74 & 0.61 & 0.85 & 1.3 & 3.41 \\

  AR \cite{yang2014depth_recovery} & 0.92 & 1.23 & 2.1 & 3.9 & 0.74 & 0.85 & 1.23 & 1.96 & 0.8 & 0.97 & 1.35 & 2.24 & 0.73 & 0.93 & 1.34 & 2.24 & 0.72 & 0.82 & 1.25 & 2.16 & 0.77 & 0.87 & 1.3 & 2.73 \\

  Ours & \textbf{0.68} & \textbf{1.07} & \textbf{1.69} & \textbf{3.01} & \textbf{0.54} & \textbf{0.8} & \textbf{1.1} & \textbf{1.65} & \textbf{0.61} & \textbf{0.85} & \textbf{1.15} & \textbf{1.78} & \textbf{0.6} & \textbf{0.83} & \textbf{1.17} & \textbf{2.01} & \textbf{0.53} & \textbf{0.76} & \textbf{1.11} & \textbf{1.74} & \textbf{0.55} & \textbf{0.82} & \textbf{1.12} & \textbf{2.13} \\

  \hline
\end{tabular}
}
\end{table*}
%---------------------------------------------------------------

%----------------------------- Figure -------------------------------
\begin{figure*}
\centering
  \includegraphics[width=1\linewidth]{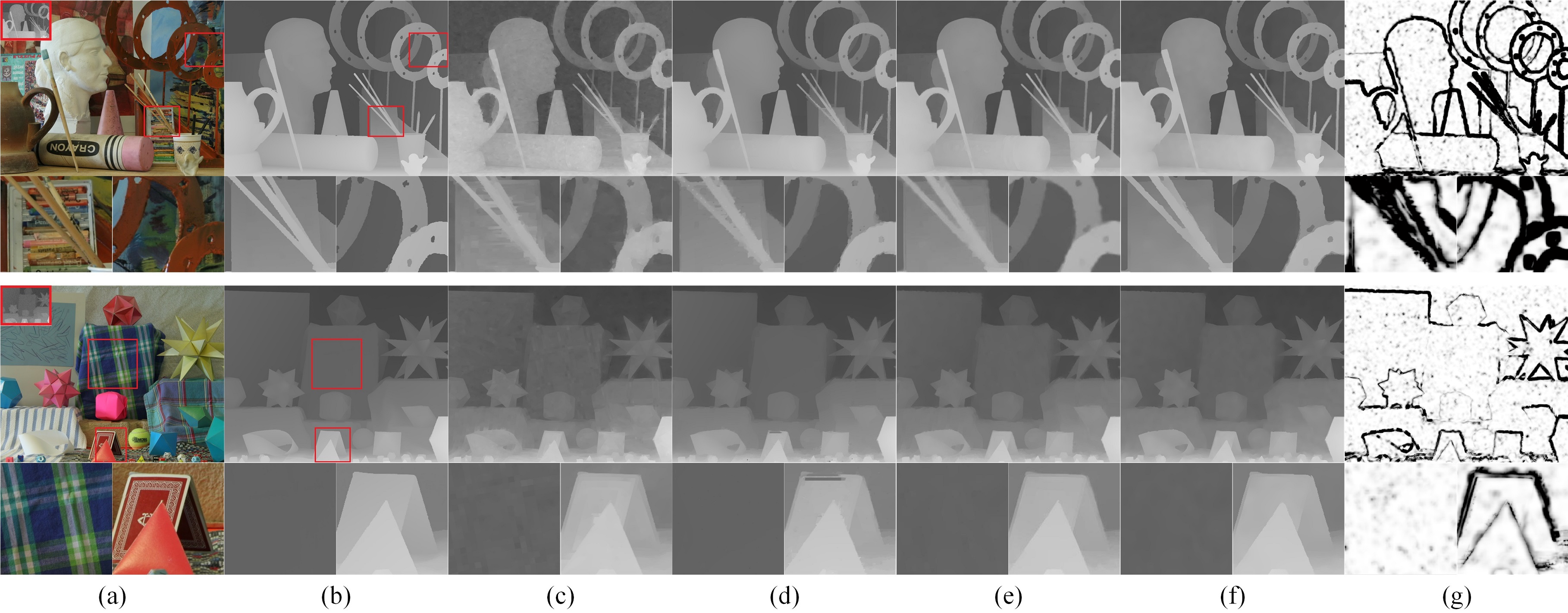}\\
  \caption{Visual comparison of $8\times$ upsampling results on simulated ToF dataset from \cite{yang2014depth_recovery}. (a) The input depth maps (in red boxes) and the corresponding color images. (b) The  groundtruth depth maps. The results of (c) the NLM-MRF in \cite{park2011high}, (d) the image guided total generalized variation upsampling in \cite{ferstl2013image}, (e) the color guided auto-regressive model \cite{yang2014depth_recovery} and (f) our method and (g) the corresponding bandwidth maps by our bandwidth selection. Regions in red boxes are highlighted.}\label{FigSimulated}
\end{figure*}
%------------------------------------------------

%----------------------------- Figure -------------------------------
\begin{figure*}
\centering
  \includegraphics[width=1\linewidth]{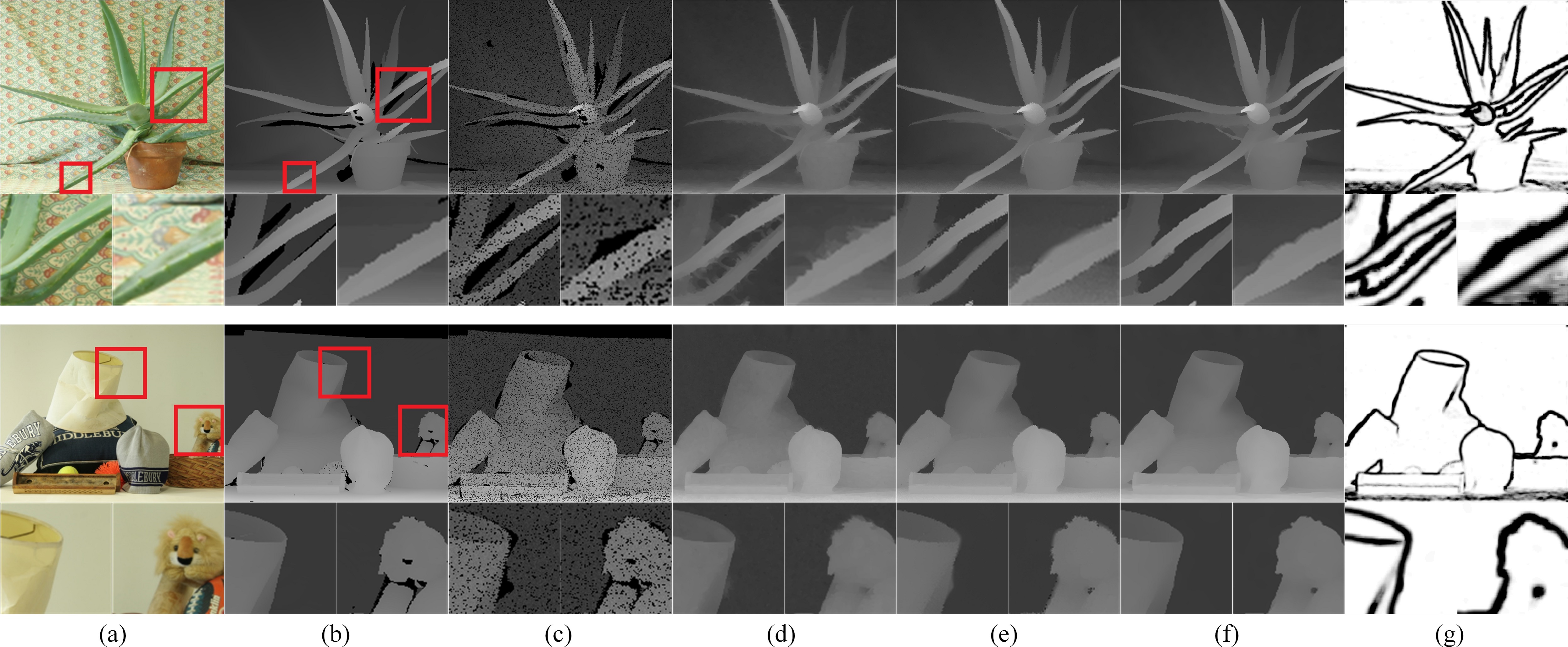}\\
  \caption{Visual comparison on simulated Kinect dataset from \cite{lu2014depth}. (a) The color images. (b) The groundtruth. (c) The degraded depth maps. The results of (d) the NLM-MRF in \cite{park2011high}, (e) the color guided auto-regressive model \cite{yang2014depth_recovery}, (f) our method and (g) the bandwidth maps by our bandwidth selection. Regions in red boxes are highlighted.}\label{FigSimulatedKinect}
\end{figure*}
%------------------------------------------------

%----------------------------- Figure -------------------------------
\begin{figure*}
\centering
  \includegraphics[width=1\linewidth]{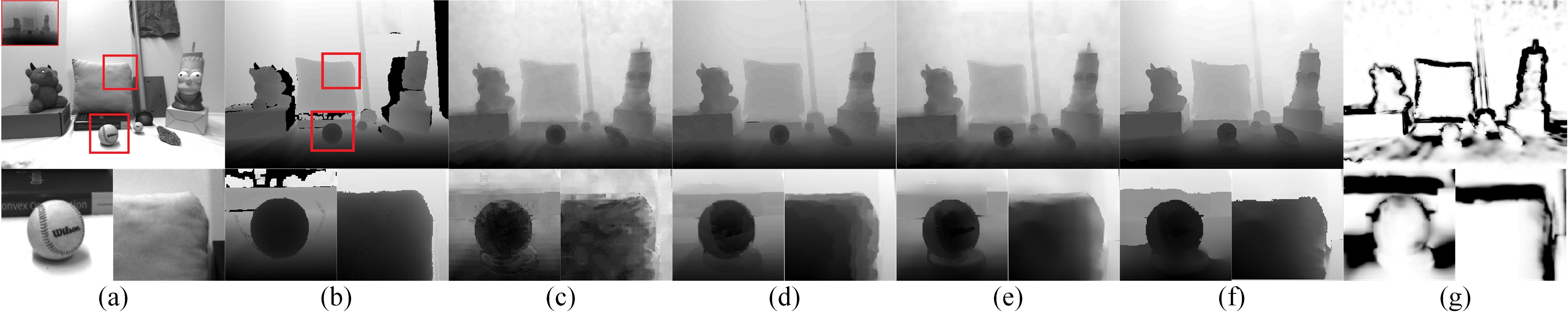}\\
  \caption{Visual comparison on real ToF dataset form \cite{ferstl2013image}. (a) The input depth map (in the red box) and the corresponding color image. (b) The groundtruth depth map. The result of (c) the NLM-MRF in \cite{park2011high}, (d) the image guided total generalized variation upsampling in \cite{ferstl2013image}, (e) the color guided auto-regressive model \cite{yang2014depth_recovery} and (f) our method and (g) the corresponding bandwidth map by our bandwidth selection. Regions in red boxes are highlighted.}\label{FigToF}
\end{figure*}
%------------------------------------------------

%----------------------------- Figure -------------------------------
\begin{figure*}
\centering
  \includegraphics[width=0.97\linewidth]{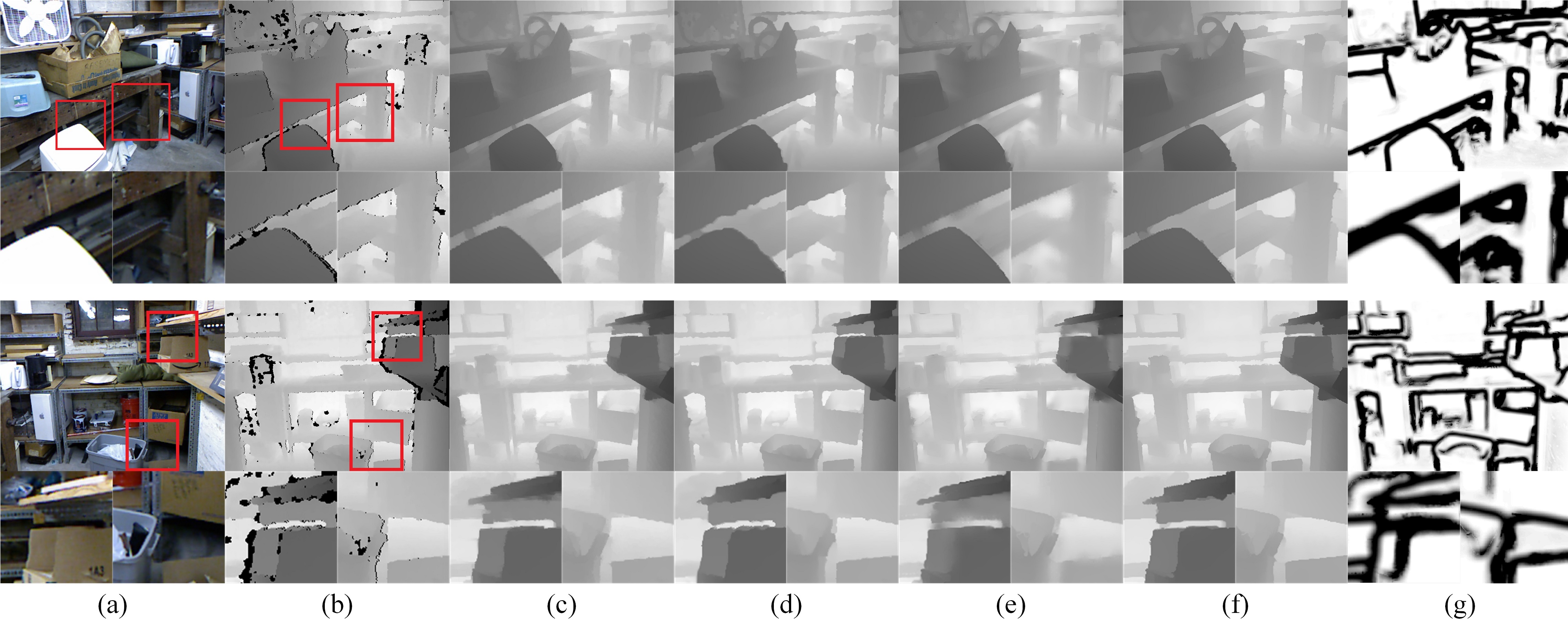}\\
  \caption{Visual comparison on NYU Kinect dataset from \cite{silberman2012indoor}. (a) The color images. (b) The RAW depth maps. The results of (c) the NLM-MRF in \cite{park2011high}, (d) joint intensity and depth co-sparse analysis model in \cite{kiechle2013joint}, (e) the color guided auto-regressive model \cite{yang2014depth_recovery}, (f) our method and (g) the corresponding bandwidth maps by our bandwidth selection.}\label{FigKinect}
\end{figure*}
%------------------------------------------------

%\subsection{Experiments on simulated datasets}
{\bf Experiments on simulated datasets}
We use the simulated ToF dataset from \cite{yang2014depth_recovery} and compare our method with the method in \cite{park2011high} which we denote as NLM-MRF, the image guided anisotropic total generalized variation upsampling in \cite{ferstl2013image}, the joint geodesic upsampling in \cite{liu2013joint}, the joint intensity and depth co-sparse analysis model in \cite{kiechle2013joint} and the color guided auto-regression upsampling in \cite{yang2014depth_recovery}\footnote{Their published MATLAB code requires huge amount of memory which is impractical for modern computers. The size of their results are also not the same as the guidance image. We thus implemented their code by ourselves with sparse matrixes and solve their normal equation with the same PCG \cite{krishnan2013efficient} as ours.}. The upsamling results are evaluated in root mean square error (RMSE) between the original depth map and the upsampling result. All the values of both the color image and the depth map are normalized into interval $[0, 1]$ for convenience. However, the RMSE is still reported in terms that all the values of the depth map are in interval $[0, 255]$. The parameters are set as follows: $\alpha$ in Eq.~(\ref{EqMyModel}) is set as $0.7/0.75/0.8/0.9$ for  $2\times /4\times /8\times /16\times $ upsampling. $\beta$ in Eq.~(\ref{EqBandwidthModel}) is set to $0.5$. The neighborhood $N(i)$ is chosen as a $9\times 9$ square patch. $\sigma _s$ and $\sigma_c $in Eq.~(\ref{EqColorSpatialWeight}) is set to $4$ and $\frac{10}{255}$ respectively. The initial value of $\lambda $ in Eq.~(\ref{EqBandwidthUpdate}) is set to $\frac{7}{255}$ for all $i\in\Omega$. Its updating rate $\tau$ in Eq.~(\ref{EqBandwidthUpdate}) is $0.3$. We also reuse the simulated Kinect dataset from \cite{lu2014depth}. The parameter setting for this dataset is the same as that of the $8\times$ upsampling in the simulated ToF upsampling.

Table \ref{TabSimulatedRMSE} summarizes the quantitative comparison in terms of RMSE. It is clear that our method outperforms all the compared methods especially for large upsampling factors $8\times$ and $16\times$. Fig.~\ref{FigSimulated} further shows visual comparison with some $8\times$ upsampling examples. As highlighted, our method can well preserve sharp depth discontinuities and suppress the texture copy artifacts where depth discontinuities are inconsistent with color edges. Especially, our method can preserve sharper depth discontinuities than the auto-regressive model \cite{yang2014depth_recovery} which only use bicubic interpolation of the input depth map. Table \ref{TabSimulatedKinectRMSE} shows the mean RMSE comparison on the simulated Kinect dataset \cite{lu2014depth}. Our method also yields lowest mean RMSE on this dataset. Fig.~\ref{FigSimulatedKinect} shows the visual comparison. Our method offers better performance in both smoothing the noise and preserving sharp depth discontinuities than other compared methods.

%\subsection{Experiments on real datasets}
{\bf Experiments on real datasets}
%
%------------------------------- Table ---------------------------------------
\begin{table}
\centering
\caption{Quantitative comparison on simulated Kinect data from \cite{lu2014depth}. Results are evaluated using the mean RMSE of all the data. The best results are in bold.}\label{TabSimulatedKinectRMSE}

\resizebox{0.43\textwidth}{!}
{
\begin{tabular}{|c|c|c|c|c|}
  \hline
   &NLM-MRF \cite{park2011high} & JID \cite{kiechle2013joint} & AR \cite{yang2014depth_recovery} & Ours \\
  \hline
  mean RMSE & 1.27 & 2.15 & 1.02 & \textbf{0.75} \\
  \hline
\end{tabular}
}
\end{table}
%
%----------------------------------------------------------------------
%
To further test the proposed method, we also perform experiments on real datasets including the real ToF data set from \cite{ferstl2013image} and the NYU Kinect dataset \cite{silberman2012indoor}. As summarized in table \ref{TabRealRMSE}, our method also achieve lowest RMSE when compared with other state-of-the-art methods for ToF depth maps upsampling. Fig.~\ref{FigToF} shows visual comparison with other methods. Our method performs better in suppressing the texture copy artifacts than others (enlarge the figure and see the 'ball' for clear comparison) and preserving sharp depth discontinuities. Fig.~\ref{FigKinect} shows the visual comparison on the Kinect data restoration. Our method has significant improvement over other methods on this dataset. Our method can properly smooth the noise, fill in the holes and rectify the jaggy depth discontinuities. Moreover, the depth discontinuities are quite sharp in our results. Results of the NLM-MRF \cite{park2011high} and color guided auto regressive model \cite{yang2014depth_recovery} suffer from blurring depth discontinuities. Results of the joint intensity and depth co-sparse analysis model \cite{kiechle2013joint} suffer from jaggy depth discontinuities.

%------------------------------- Table ---------------------------------------
\begin{table}
\centering
\caption{Quantitative comparison on real ToF dataset from \cite{ferstl2013image}. The error is calculated as RMSE to the measured groundtruth in mm. The best results are in bold.}\label{TabRealRMSE}

\resizebox{0.5\textwidth}{!}
{
\begin{tabular}{|c|cccccc|}
  \hline
   & JGF \cite{liu2013joint} & JBF \cite{kopf2007joint} & NLM-MRF \cite{park2011high} & TGV \cite{ferstl2013image} & AR \cite{yang2014depth_recovery} & Ours \\
  \hline
  \emph{Books} &  17.39mm & 15.42mm & 14.31mm & 12.8mm & 13.28mm & \textbf{12.33mm} \\
  \emph{Devil} &  19.02mm & 16.47mm & 15.36mm & 14.97mm & 14.73mm & \textbf{14.12mm} \\
  \emph{Shark} &  18.17mm & 17.16mm & 15.88mm & 15.53mm & 15.86mm & \textbf{14.71mm} \\
  \hline
\end{tabular}
}
\end{table}

%----------------------------------------------------------------------

%-----------------------------
\section{Conclusion}
\label{SecConclusion}
%{\bf Conclusion}
In this paper, we propose a robust data driven optimization framework for high quality image guided depth map restoration. The newly proposed data term is robust against the noise which has been validated through both mathematical analysis and experimental results. The smoothness term performs well in suppress texture copy artifacts and performs especially better in preserving sharp depth discontinuities than other state-of-the-art methods. Given that our model is highly non-convex and directly solving it is challenging, we propose a numeric solution which can be efficiently solved and easily implemented by solving linear systems using off-the-shelf linear system solvers. The proposed data driven bandwidth selection further helps to preserve sharper depth discontinuities and fine details even for large upsampling factors. Experimental results on both simulated data and real data have shown our method outperforms other state-of-the-art methods.

{\small
\bibliographystyle{ieee}
\bibliography{egbib}
}

\end{document}